# Modeling and Force Analysis of a Soft Linear Pneumatic Actuator

By Mohammed Abboodi mabbo103@uottawa.ca

## Abstract

Soft sleeve actuators (SSAs) have recently been developed as a pneumatic actuation approach for wearable and assistive robotic systems. By integrating the actuation structure into a sleeve-like geometry, these actuators can reduce reliance on external attachment layers and transmission mechanisms while maintaining compliance with limb-shaped surfaces. However, the force-generation behavior of SSAs remains insufficiently explained, particularly with respect to the variation of output force during extension, the influence of external loading, and the mechanical role of axial stiffness. This paper presents an analytical and experimental force analysis of a linear soft sleeve actuator (LSSA). A quasi-static analytical model was developed by expressing the net axial force as the pressure-generated contribution from the cap and folded walls, reduced by the force associated with axial stiffness. The model incorporates internal pressure, projected pressure areas, folded wall geometry, axial displacement, and an experimentally fitted axial stiffness relation. Prescribed-extension and static-load experiments were conducted to evaluate the actuator response. At 125 kPa, the generated force decreased from approximately 112 N at zero extension to nearly zero at 40 mm. Static loading delayed measurable force generation and reduced force output, particularly at low and intermediate pressures. The results show that LSSA force generation is governed by coupled effects of pressure, geometry, displacement, loading, and axial stiffness.

## 1. Introduction

Soft robots and wearable assistive devices require actuators that can mechanical out put while remaining compliant with the surrounding environment or the human body[1]. Soft pneumatic actuators meet many of these requirements by converting internal pressure into controlled deformation of flexible structures[1], [2]. Their compliance, large deformation capacity, lightweight construction, and relatively high force-to-weight ratio make them suitable for soft manipulators, rehabilitation devices, wearable robots, and other systems that require safe physical interaction [3], [4]. However, the force generated by a soft pneumatic actuator is difficult to predict because it depends not only on the applied pressure but also on the actuator geometry, material deformation, structural constraints, and loading conditions[4], [5].

Several actuation technologies have been investigated for soft and wearable robotic systems, including shape memory alloy actuators, twisted and coiled polymer actuators, electroactive polymer actuators, cable-driven systems, and pneumatic actuators. Although these technologies can provide compliant motion, each has limitations that affect practical integration. Shape memory alloy and twisted polymer actuators are often limited by thermal response and heat management [6] . Electroactive polymer actuators may require high voltage or provide limited force output [7] . Cable-driven systems can generate large forces but often require transmission components, routing paths, and attachment interfaces that may reduce comfort and increase mechanical losses. Pneumatic actuation remains one of the most widely used approaches because it can produce large deformation and useful force through a compliant structure [8]. Nevertheless, pneumatic actuators also require careful mechanical design to direct pressure-induced deformation into useful force rather than undesired radial expansion or internal structural deformation.

This issue is particularly important for wearable and assistive applications, where the actuator must deliver force over a required displacement range. Many soft pneumatic actuators are characterized using blocked-force measurements, which are useful for identifying the maximum force generated when motion is constrained [8]. However, blocked force alone does not describe how much force remains available as the actuator moves. During deformation, the effective pressure area, internal geometry, and structural resistance of the actuator can change substantially. Consequently, an actuator that generates high force at a constrained position may produce much lower force near the end of its stroke. A displacement-dependent force analysis is therefore needed to evaluate the useful operating range of soft pneumatic actuators.

Soft sleeve actuators (SSAs) have been introduced as a sleeve-based pneumatic architecture intended to reduce dependence on external attachment layers, straps, and transmission mechanisms. In this architecture, the actuator structure is integrated into a sleeve-like form that can conform to limb-shaped geometries while producing controlled motion. SSA designs have been developed for linear [8], bending [8], twisting [4] , and omnidirectional motions [5]. Among these designs, the linear soft sleeve actuator (LSSA) generates axial extension through a folded wall geometry. The actuator consists of internal and external folded walls connected by tie-restraining layers, which help guide axial deformation and reduce undesired radial expansion during pressurization [4], [5], [8].

Although the LSSA can generate axial motion and force, its force-generation behavior is governed by a coupled interaction between pressure, geometry, and axial stiffness. Internal pressure acts on the cap and folded walls to produce extension, while the actuator structure resists deformation through the force associated with axial stiffness. As the actuator extends, the folded wall geometry changes, altering the effective projected area that contributes to axial force generation. At the same time, the force associated with axial stiffness increases with displacement. This interaction reduces the net output force during extension and creates a direct relationship between actuator displacement and available force. A clear analytical explanation of this behavior is needed to interpret the LSSA response and support future design optimization.

This paper presents an analytical and experimental force analysis of the LSSA. The analytical model expresses the actuator output force as the pressure-generated contribution from the cap and folded walls, reduced by the force associated with axial stiffness. The model incorporates internal pressure, projected pressure areas, folded wall geometry, axial displacement, and an experimentally determined axial stiffness relation. The printed thermoplastic polyurethane used to fabricate the actuator is characterized through tensile testing and hyperelastic curve fitting. The actuator is then evaluated using two experimental configurations: a prescribed-extension test that measures force at different displacement positions and a static-load test that examines the effect of external axial loading on force generation.

The main contribution of this study is a force-balance analysis that explains how pressure, folded wall geometry, displacement, and axial stiffness determine the net force generated by the LSSA. The analysis clarifies why the actuator force decreases during extension, why the force approaches zero near the effective extension limit, and how external static loading modifies the force–pressure response. This work provides a basis for evaluating sleeve-based soft pneumatic actuators beyond blocked-force measurements and supports future improvement of LSSA designs through the coordinated adjustment of folded wall geometry, projected pressure area, tie-restraining layers, and axial stiffness.

The remainder of this paper is organized as follows. Section 2 describes the actuator material, experimental setups, and analytical force model. Section 3 presents the force–displacement and static-load results. Section 4 discusses the results in relation to the analytical model. Section 5 summarizes the main conclusions.

## 2. Material and Method

### 2.1 Material

Thermoplastic polyurethane with a Shore hardness of 85A (TPU 85, NinjaTek, USA) was used to fabricate the soft actuator. TPU was selected because it can undergo large elastic deformation and recover its original shape after unloading, making it suitable for soft pneumatic actuators subjected to repeated inflation and deformation. The material also exhibits a nonlinear stress-strain response at large strains.

To characterize the mechanical response of the printed TPU, tensile specimens were fabricated using the same printing conditions used for actuator fabrication. This ensured that the measured material behavior reflected the properties of the printed structure rather than those of the raw filament alone. The specimens were prepared according to the ASTM D638 standard geometry, as shown in Fig. 1(c). Tensile testing was performed using an Instron universal testing machine (Model 4202), as shown in Fig. 1(a). Each specimen was stretched under ambient conditions at a constant crosshead speed of 100 mm/min until approximately 660% elongation. The resulting stress–strain response is shown in Fig. 1(b).

The experimental stress–strain data were fitted using hyperelastic material models in ANSYS Workbench 2022. Several commonly used models for elastomeric materials were evaluated, including the Neo-Hookean, Mooney–Rivlin, Yeoh, Ogden, and Arruda–Boyce models. Among these models, the five-parameter Mooney–Rivlin model provided the most stable and accurate fit to the TPU 85 data. Therefore, this model was selected to represent the nonlinear elastic behavior of the printed TPU in the present analysis. The corresponding curve fitting is shown in Fig. 1(d), and the identified material constants are listed in Table 1.

**Table 1**: Lists the hyperelastic material constants identified for TPU 85

| Material Model | Parameters | Value (MPa) |
|---|---|---|
| Mooney-Rivlin 5-parameter | C10 | -3.1992 |
| | C01 | 6.977 |
| | C20 | 0.0281 |
| | C11 | -0.074972 |
| | C02 | 0.92155 |
| | Incompressible parameter D1 | 0 |

For the five-parameter Mooney–Rivlin model, the strain-energy density function is expressed as

$$W = C_{10}(I_1 - 3) + C_{01}(I_2 - 3) + C_{11}(I_1 - 3)(I_2 - 3) + C_{10}(I_1 - 3) + C_{20}(I_1 - 3)^2 + C_{02}(I_1 - 3)^2 \quad (1)$$

where $W$ is the strain-energy density, $I_1$ and $I_2$ are the first and second strain invariants, and $C_{10}$, $C_{01}$, $C_{20}$, $C_{11}$, and $C_{02}$ are material constants obtained from curve fitting. The material was treated as incompressible during the fitting process, with the incompressibility parameter $D_1$ set to zero. The fitted constants for TPU 85 are reported in Table 1.

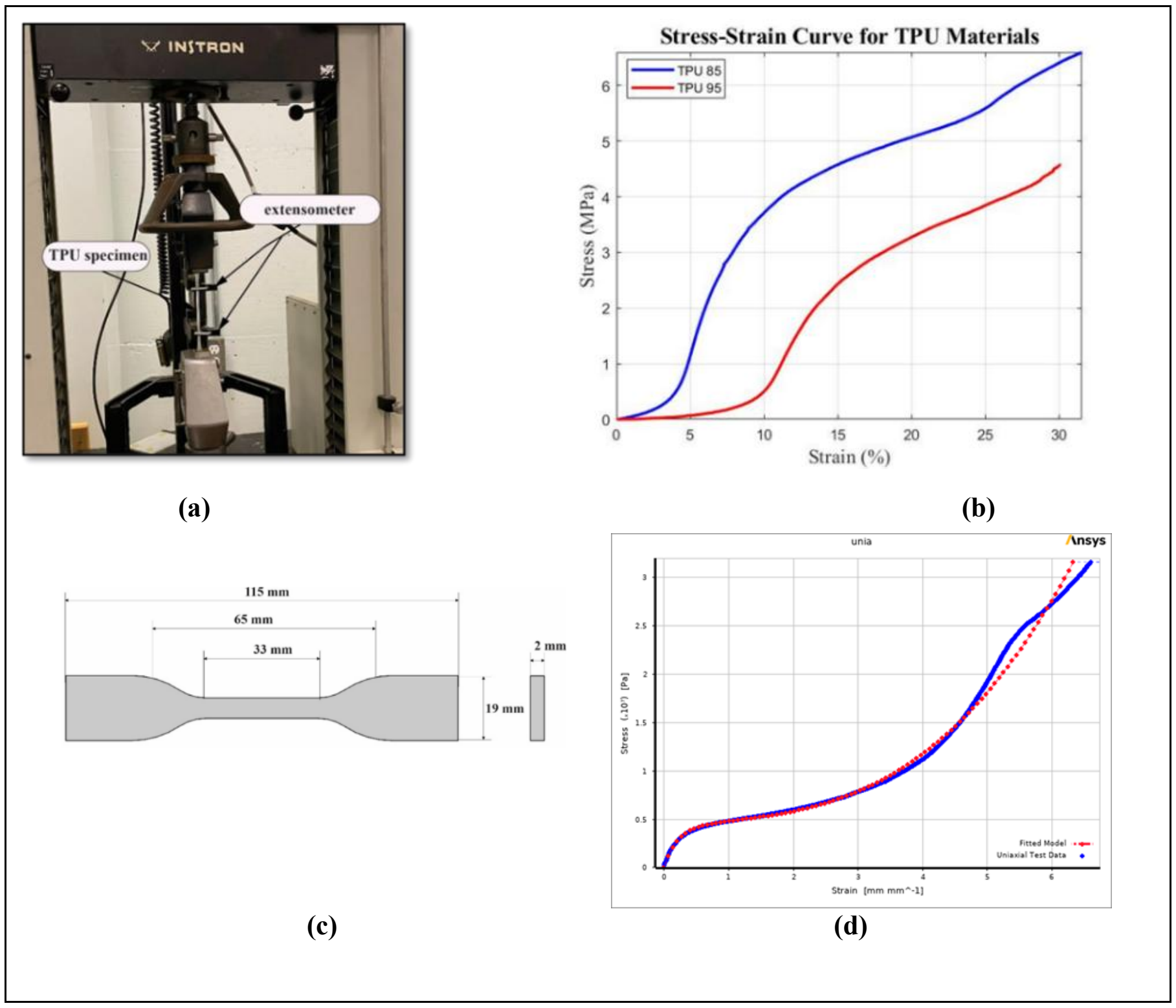

**Figure 1:** (a) Tensile test experimental setup (b) stress-strain curve (c) specimen design (d) Curve fitting of TPU material using Mooney Rivlin's 5-parameter material

## 2.2 Experimental Setup

Two experimental setups were used to evaluate the force-generation behavior of the LSSA. The first setup measured the actuator output force at prescribed displacement positions during extension, as shown in Fig. 2 (a). In this setup, the LSSA was mounted on an Instron testing machine using a custom actuator fixture that maintained axial alignment between the actuator and the loading direction. The actuator was pressurized using a proportional pressure regulator (QBX pressure control valve, Proportion-Air, Inc.), which maintained the selected pressure levels during testing. This setup allowed the force response of the actuator to be measured at different extension positions under controlled pressure conditions.

The second setup was used to examine the effect of external axial loading on the force generated by the LSSA, as shown in Fig. 2(b). A stainless-steel cable transmitted the applied load to the proximal end of the actuator through a coupling unit connected to the force measurement assembly. The actuator was fixed at one end using a mounting module that provided stable support while allowing the required axial motion during loading.

Axial force was measured using a TE Connectivity FC223 load cell (FC22310050). The load cell was housed in a purpose-designed 3D-printed PLA fixture that aligned the sensing axis with the actuator axis. The fixture provided three-axis adjustment to improve collinearity among the actuator, cable, and load cell and to reduce off-axis loading. The load cell operated with a 5 V excitation voltage and had an accuracy of 1% across the tested force range.

Pressure and force signals were collected using a National Instruments PXIe-1073 data acquisition system connected to a LabVIEW-based graphical interface. The system synchronized sensor readings, recorded the experimental data, and provided real-time visualization during testing. Together, these two setups allowed the LSSA force response to be characterized under prescribed-displacement and external-loading conditions.

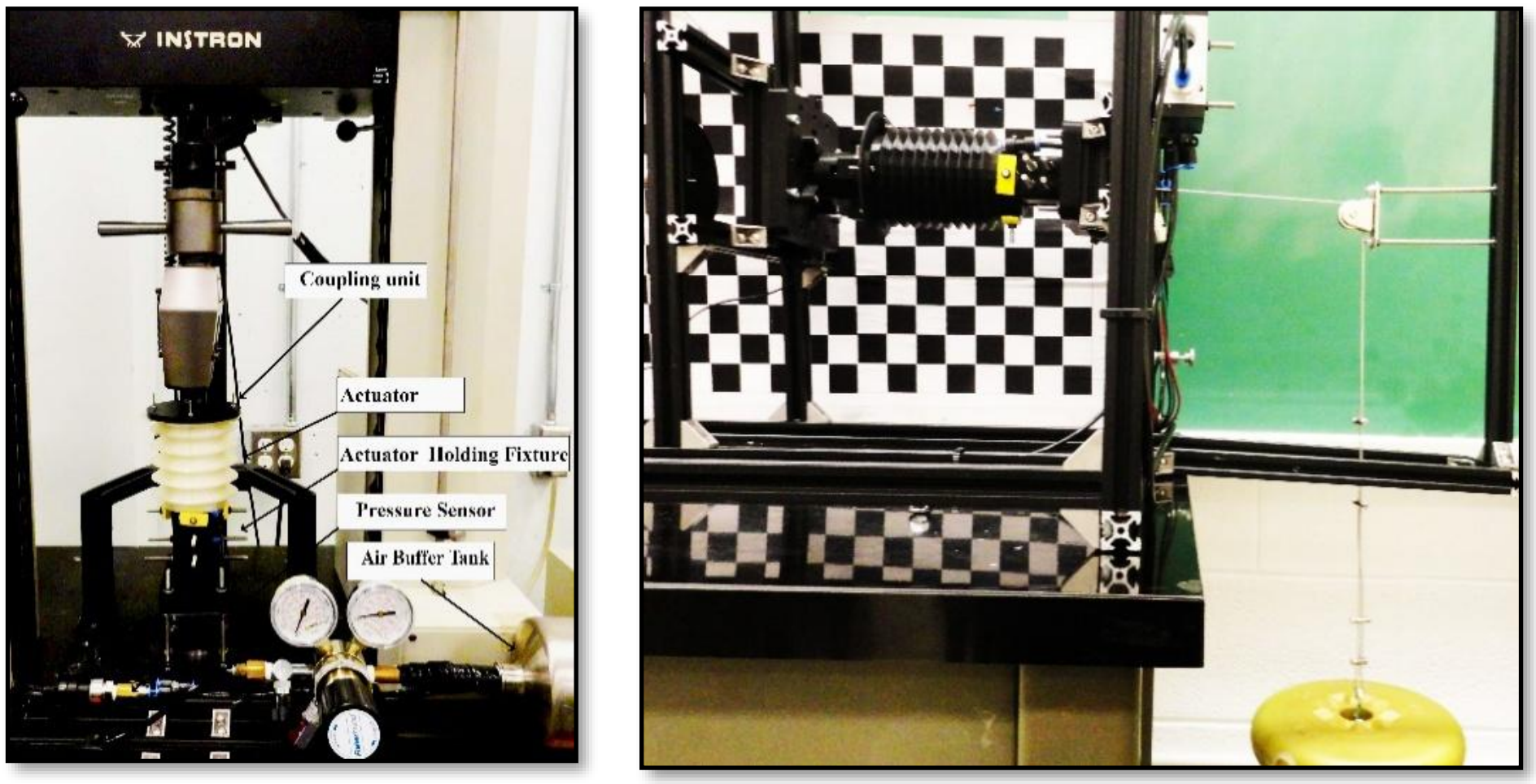

(a) (b)

**Figure 2:** Force-characterization setups: (a) prescribed-extension test (b) static-load test.

## 2.3 Analytical Force Model

An analytical model was developed to estimate the axial force generated by the linear soft sleeve actuator (LSSA). The model relates the actuator output force $Fy$ to the internal pressure $P$, actuator geometry, and axial displacement y. The analysis was conducted under quasi-static conditions; therefore, dynamic effects such as inertia and damping were neglected. The internal pressure was also assumed to be uniformly distributed within the actuator.

When the LSSA is pressurized, it extends in the axial direction until its motion is constrained by an external boundary, a prescribed displacement, or an applied load. At equilibrium, the actuator output force is determined by the pressure-induced forces acting on the cap and folded walls, reduced by the force associated with the axial stiffness of the actuator. The force balance is expressed as

$$Fy = \sum (F_1 + F_{2y} - F_{3y}) - F_K \quad \textbf{(2)}$$

Where $F_1$ is the force generated by the cap, $F_{2y}$ is the axial component of the force generated by the external wall, $F_{3y}$ is the axial component of the force generated by the internal wall, and $F_K$ is the force associated with the axial stiffness of the actuator. The geometric parameters and force components used in the model are shown in Fig. 3(a), and the detailed wall geometry is shown in Fig. 3(b).

The pressure-induced force components are calculated from the internal pressure and the corresponding projected areas:

$$F_1 = P \times A_1 \quad \textbf{(3)}$$

$$F_{2y} = P \times A_2 \quad \textbf{(4)}$$

$$F_{3y} = P \times A_3 \quad \textbf{(5)}$$

where $A_1$, $A_2$, and $A_3$are the projected areas of the cap, external wall, and internal wall, respectively. The cap is treated as an annular surface with outer radius $R_{1o}$and inner radius $R_{1i}$. Therefore, the projected area of the cap is

$$A_1 = \pi \times (R_{1o}^2 - R_{1i}^2) \quad \textbf{(6)}$$

The axial force generated by the external wall results from the internal pressure acting on the effective projected area of the folded geometry. Considering the hollow structure of the actuator, this area is calculated as the difference between two concentric circular areas:

$$A_2 = \pi(R_{2i} + S \times \cos(\theta))^2 - \pi\,(R_{2i})^2 \quad \textbf{(7)}$$

where $R_{2i}$ is the inner radius of the external wall, $S$ is the fold length, and $\theta$ is the fold angle relative to the horizontal direction, as defined in Fig. 3(b). Similarly, the projected area associated with the internal wall is expressed as

$$A_3 = \pi(R_{3i} + S\cos\theta)^2 - \pi R_{3i}^2 \quad \textbf{(8)}$$

where $R_{3i}$ is the corresponding radius of the internal wall, as defined in Fig. 3(b). The term $F_{3y}$ is subtracted in Eq. (2) because its axial component acts opposite to the extension force generated by the cap and external wall.

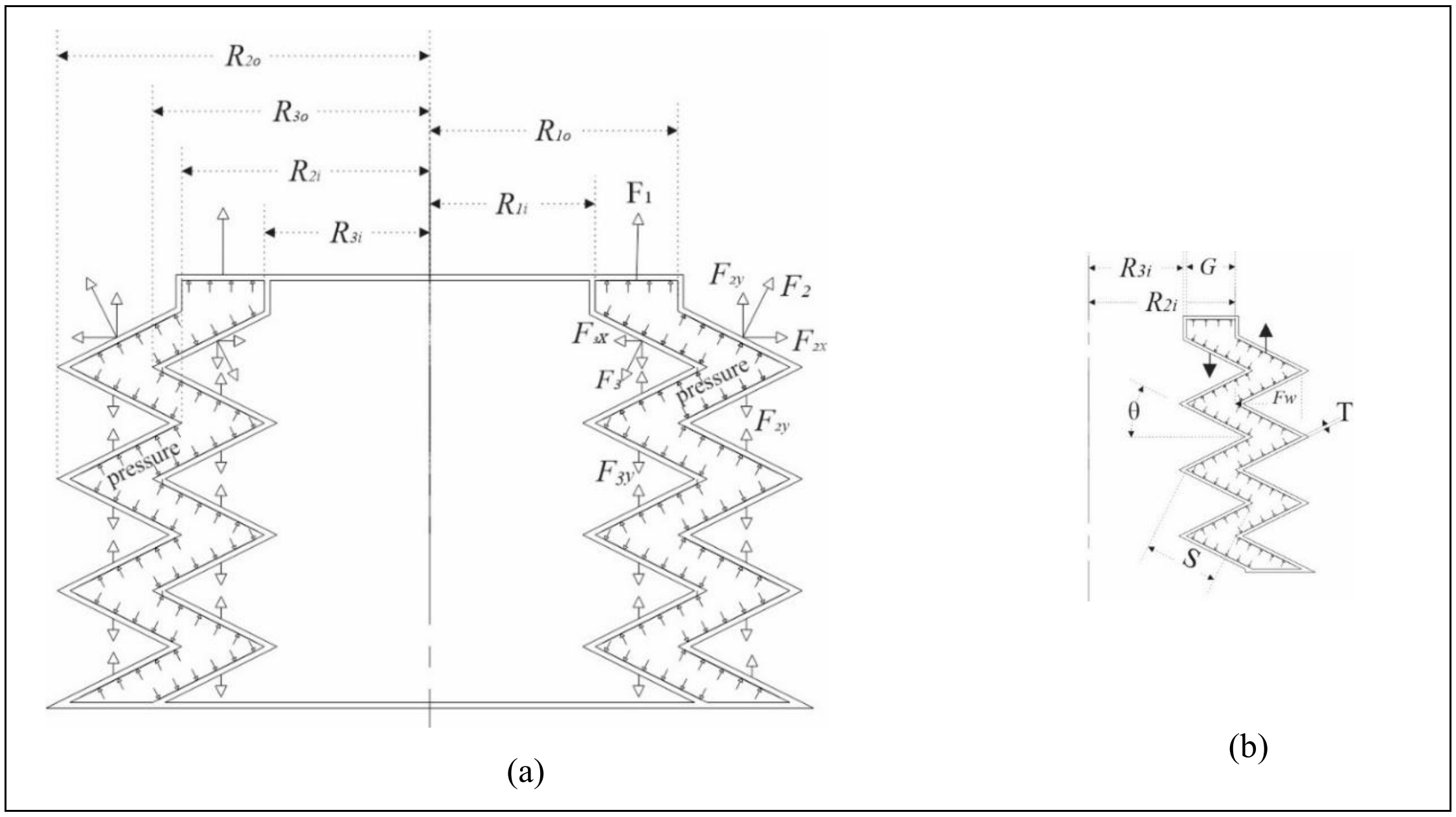

**Figure 3:** (a) Geometric representation of the LSSA and the forces acting on it, (b) Detailed wall geometrical parameters

The actuator also resists extension because of its axial stiffness, which results from the combined effects of material elasticity and geometric constraints. The force associated with the axial stiffness was obtained experimentally and represented using the polynomial fit shown in Fig. 4:

$$FK_{actuator} = a\,y^3 + b\ y^2 + c\,y + d \quad \textbf{(9)}$$

where $a$, $b$, $c$, and $d$are empirical constants determined from the axial stiffness experiment. For the LSSA actuator model, the fitted polynomial is

The axial stiffness of the actuator, $K_{\text{axial}}$, is obtained by differentiating $F_K$with respect to the axial displacement $y$:

$$FK_{actuator} = 4.1481 \times 10^{-4}\,y^3 + 1.2865 \times 10^{-2}\ y^2 + 2.0789\,y - 0.2246 \quad \textbf{(10)}$$

$$K_{axial} = \frac{d}{dy} FK_{actuator} \quad \textbf{(11)}$$

$$K_{axial} = 1.24443 \times 10^{-3} y^2 + 2.5730 \times 10^{-2} y + 2.0789 \quad \textbf{(12)}$$

By substituting Eq. (10) into Eq. (2), the static force model of the LSSA becomes

$$Fy = F_1 + F_{2y} - F_{3y} - 4.1481 \times 10^{-4} y^3 - 1.2865 \times 10^{-2} y^2 - 2.0789 y + 0.2246 \quad \textbf{(13)}$$

or, in terms of the projected areas,

$$Fy = P(A_1 + A_2 - A_3) - 4.1481 \times 10^{-4} y^3 - 1.2865 \times 10^{-2} y^2 - 2.0789 y + 0.2246 \quad \textbf{(14)}$$

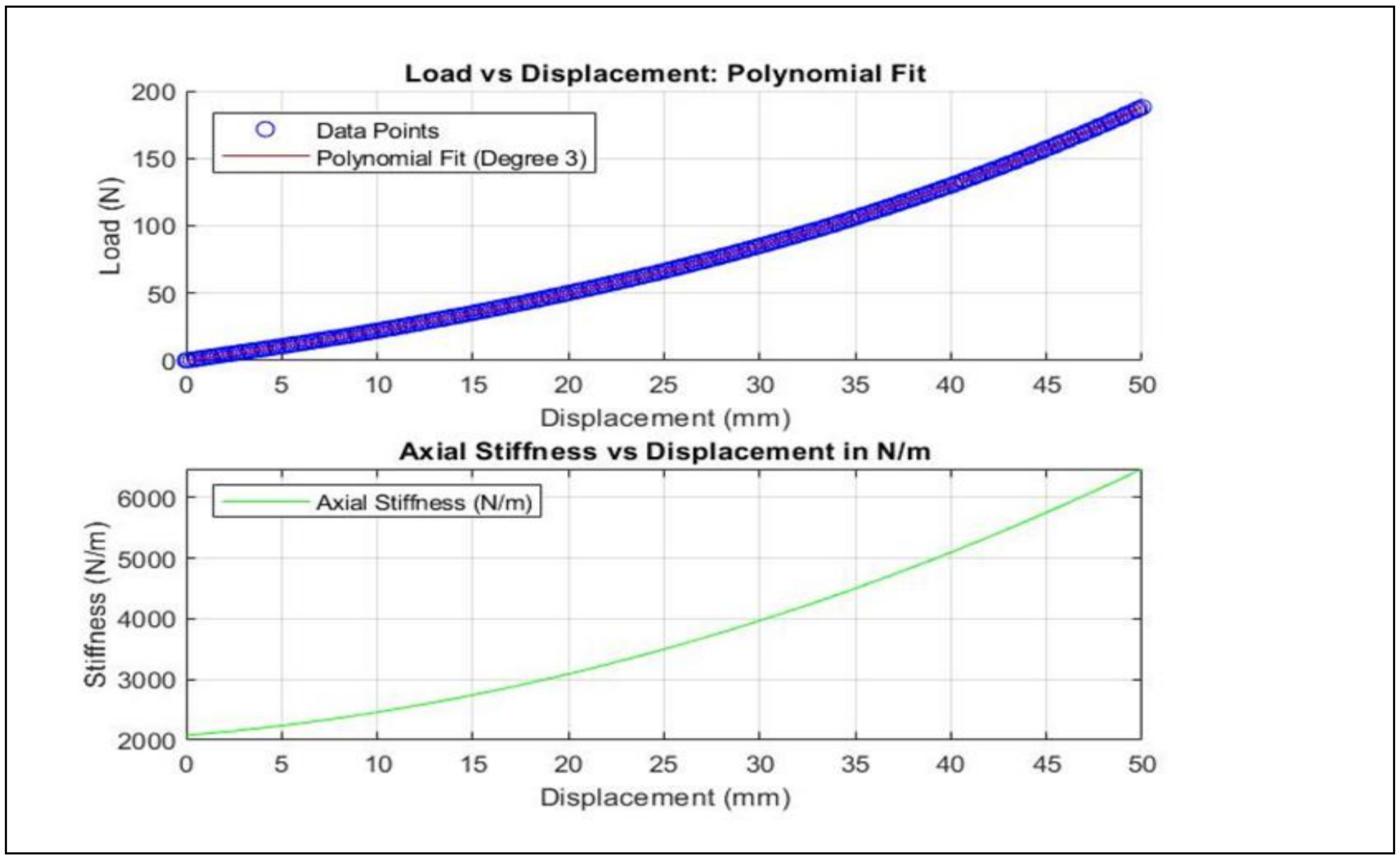

**Figure 4:** Axial stiffness polynomial model

This formulation separates the force generated by internal pressure from the force required to overcome the actuator's axial stiffness. The pressure term represents the contribution of the actuator geometry, while the axial stiffness term accounts for the structural resistance during extension. The model was used to predict the LSSA output force at prescribed displacement positions and to compare the analytical predictions with the experimental measurements.

## 3. Results

### 3.1 Force Generated at Prescribed Extension Positions

The force–displacement response of the LSSA was evaluated at a constant internal pressure of 125 kPa using the prescribed-extension setup described in Section 2.2 and shown in Fig. 2(a). During the test, the actuator was mounted on the Instron testing machine, pressurized to the target pressure, and gradually extended by increasing the crosshead displacement. The generated axial force was recorded at successive extension positions until the actuator reached its maximum extension.

The measured force–displacement response is shown in Fig. 5. The LSSA generated its maximum force at the initial position, with an output of approximately 112 N at zero extension. As the extension increased, the generated force decreased continuously, reaching approximately 72 N at 10 mm, 41 N at 20 mm, and 19 N at 30 mm. At approximately 40 mm of extension, the generated force approached zero, indicating that the actuator had reached its effective extension limit under the tested pressure condition.

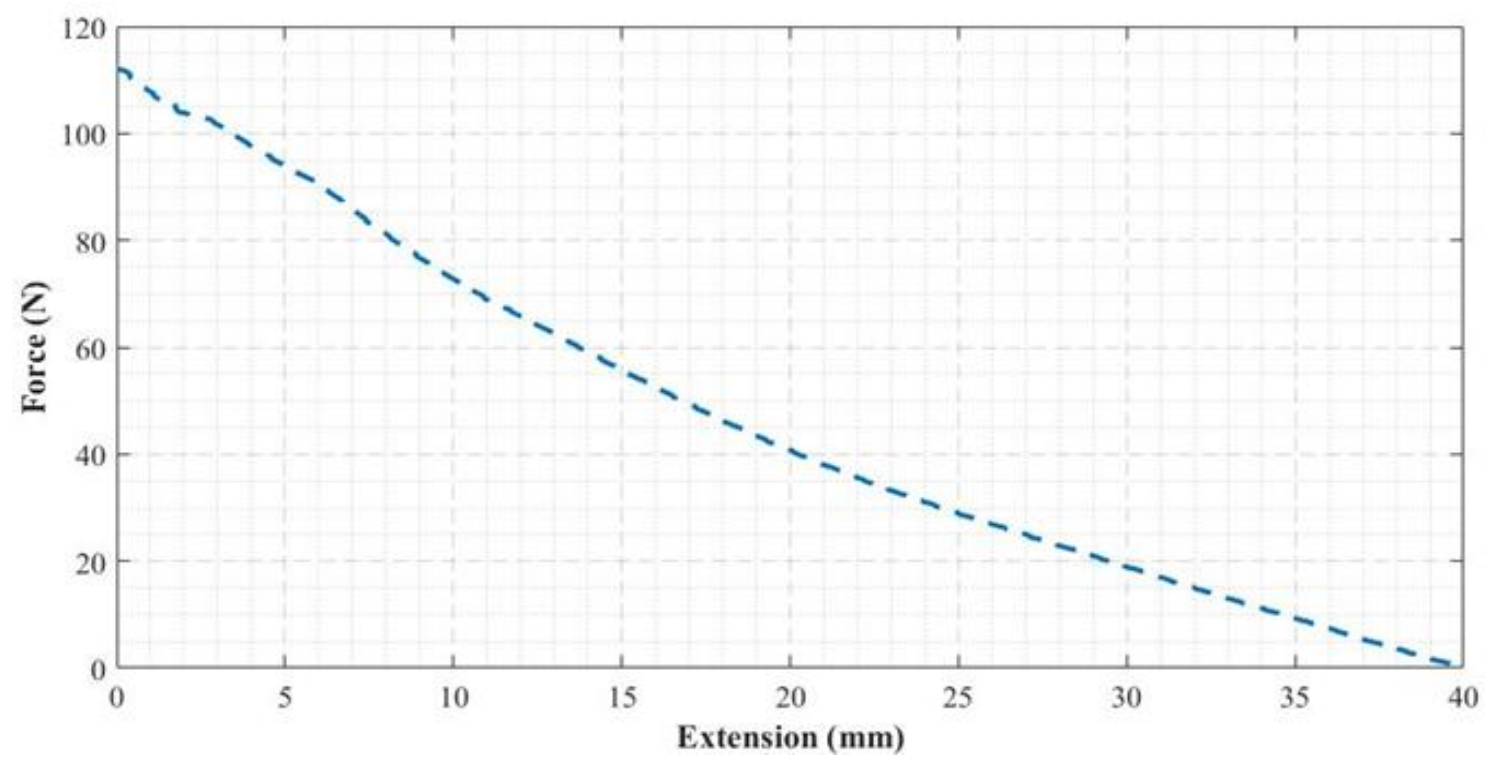

**Figure 5:** Force–displacement response of the LSSA

The curve shows a nonlinear reduction in force over the extension range. The largest force reduction occurred during the early stage of extension, followed by a more gradual decline as the actuator approached its maximum extension. These results show that the available axial force depends strongly on the actuator extension state, with the highest force generated near the constrained position and the lowest force generated near full extension.

### 3.2 Effect of Static Load on Force Generation

The effect of external static loading on LSSA force generation was evaluated using the static-load setup described in Section 2.2 and shown in Fig. 2(b). The actuator was tested under four loading conditions: 0 kg, 1 kg, 2 kg, and 3.5 kg. The 0 kg condition was used as a no-load reference. For each condition, the generated force was recorded as the internal pressure increased from 0 to 200 kPa.

The force–pressure responses under the four static-load conditions are shown in Fig. 6. Under the no-load condition, the actuator force increased continuously with pressure, reaching approximately 160 N at 200 kPa. With a 1 kg load, the force followed a similar trend but remained slightly lower than the no-load response across most of the pressure range, reaching approximately 155 N at 200 kPa. The 2 kg load produced a greater reduction in force at low and intermediate pressures; however, its response became closer to the lower-load cases at higher pressures.

The largest reduction was observed under the 3.5 kg load. In this case, the measured force remained near zero until approximately 60–70 kPa, after which it increased steadily with pressure and reached approximately 130 N at 200 kPa. This response shows that higher external loads delayed the onset of measurable force generation and reduced the force output over all the tested pressure range.

Overall, the results show that the LSSA maintained pressure-dependent force generation under all tested loading conditions. However, increasing the external static load reduced the measured force output, with the effect being most pronounced at low and intermediate pressures.

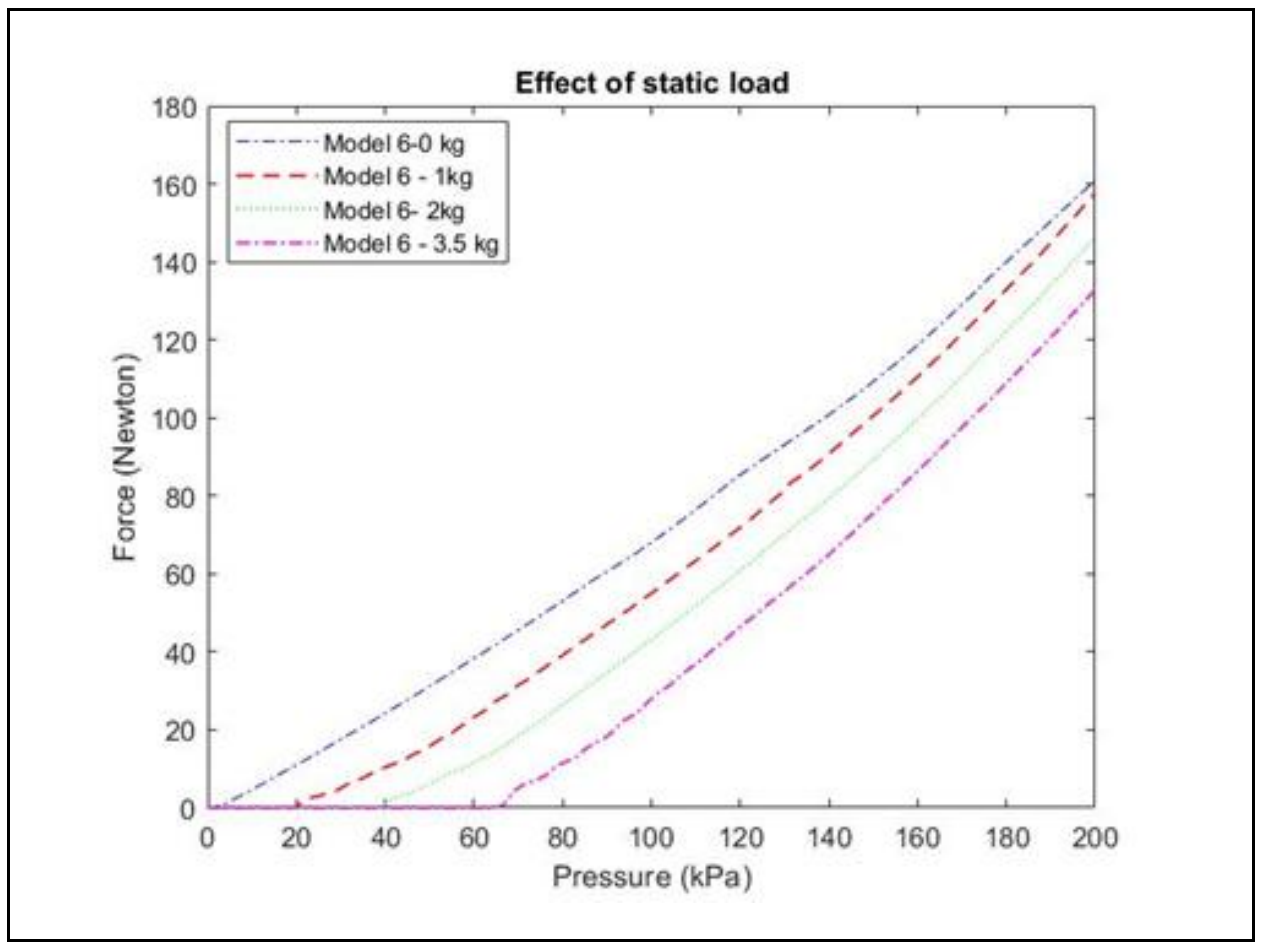

**Figure 6:** Effect of static load on LSSA force generation during extension.

## 4. Discussion

The force-generation behavior of the linear soft sleeve actuator (LSSA) is governed by the balance between pressure-induced actuation and structural resistance. The analytical model expresses this balance by separating the pressure-generated force from the force associated with axial stiffness. This formulation is central to interpreting the experimental results because it shows that the LSSA cannot be treated as a simple pneumatic chamber with a constant effective area. Instead, its output force depends on the changing folded wall geometry, the projected pressure areas, and the displacement-dependent resistance of the actuator structure.

The force–displacement results are consistent with this interpretation. As the LSSA extends, the folded wall geometry changes, reducing the effective projected area that contributes to axial force generation. At the same time, the force associated with axial stiffness increases with displacement and progressively opposes the pressure-generated force. As a result, the net output force decreases as extension increases. The approach to zero force near the end of the stroke represents an equilibrium condition in which the remaining pressure-generated axial force is balanced by the force required to overcome axial stiffness. This condition defines the effective extension limit of the actuator under the tested pressure condition.

This behavior highlights an important design feature of the LSSA: force and displacement are coupled. The actuator can generate high force near the constrained position, but the available force decreases as it approaches full extension. Therefore, blocked force alone is not sufficient to evaluate LSSA performance. For practical applications, especially wearable or assistive systems, the useful force should be assessed over the required displacement range rather than only at the initial position.

The static-load experiments further support the force-balance interpretation. Applying an external load adds resistance that must be overcome before measurable output force is produced. This effect is most evident at low and intermediate pressures, where the pressure-generated force is still limited. As pressure increases, the actuator produces sufficient force to overcome a greater portion of the combined resistance from axial stiffness and external loading. Thus, the static load shifts the force response downward and delays force generation, but it does not change the basic pressure-driven actuation mechanism.

The analytical model helps clarify the design variables that control this behavior. The pressure term depends on the projected areas of the cap and folded walls, whereas the axial stiffness term represents the resistance introduced by the material, wall geometry, and tie-restraining layers. Increasing the effective projected area can improve force generation, whereas excessive axial stiffness reduces the net force available for external work. This creates a design trade-off: the actuator must be sufficiently stiff to preserve its sleeve geometry and prevent uncontrolled deformation, but sufficiently compliant to allow efficient extension and force transmission.

The role of axial stiffness is therefore not purely limiting. It contributes to structural stability and helps maintain the relative position of the internal and external walls during pressurization. However, when axial stiffness becomes too high, a larger portion of the pressure-generated force is consumed internally, reducing the output force delivered by the actuator. Optimizing the folded wall geometry and tie-restraining layers is therefore essential for improving the useful force range of the LSSA.

The proposed model provides a practical framework for interpreting LSSA force generation, but it remains quasi-static. It neglects inertia, damping, viscoelastic hysteresis, transient pressure losses, and time-dependent material effects. These assumptions are appropriate for the prescribed-extension and static-load tests used in this study, but they limit the model's application to high-speed or cyclic actuation. In addition, the axial stiffness term was obtained from an experimentally fitted polynomial, meaning that recalibration may be required for other actuator geometries, materials, or manufacturing conditions.

## 5. Conclusion

This study presented an analytical and experimental force analysis of a LSSA. The proposed force-balance model described the actuator output force as the pressure-generated contribution from the cap and folded walls, reduced by the force associated with axial stiffness. The experimental results confirmed that the LSSA force output is displacement-dependent and decreases as the actuator approaches its effective extension limit. Static-load testing further showed that external loading reduces the available force and delays measurable force generation, particularly at low and intermediate pressures.

These findings show that LSSA force generation is governed by the coupled effects of internal pressure, folded wall geometry, axial displacement, external loading, and axial stiffness. The model provides a basis for interpreting the actuator's force response and guiding future design refinement. Future work will focus on developing a numerical model of the LSSA, comparing its predictions with the analytical model and experimental results, and extending the analysis to dynamic operation, cyclic loading, and material hysteresis.